\documentclass[11pt]{article}

\usepackage[utf8]{inputenc}
\usepackage[T1]{fontenc}
\usepackage{amsmath,amssymb}
\usepackage{booktabs}
\usepackage{graphicx}
\usepackage[margin=1in]{geometry}
\usepackage{authblk}
\usepackage{float}
\usepackage{tikz}
\usetikzlibrary{arrows.meta, positioning, shapes.geometric}
\usepackage[colorlinks=true,linkcolor=blue,citecolor=blue,urlcolor=blue]{hyperref}

\title{SparPR: Runtime KV-Cache Recovery for Resource-Constrained LLM Inference on Consumer Hardware}
\author[1]{Jieyun Deng}
\author[2]{Keming Hu}
\author[3]{Yanghe Liu}
\author[1]{Li Feng\thanks{Corresponding author. E-mail: fengli\_xjtu@163.com}}
\affil[1]{Southwest Jiaotong University, Chengdu, China}
\affil[2]{Wuhan Institute of Technology, Wuhan, China}
\affil[3]{Southwest Petroleum University, Chengdu, China}
\date{}

\begin{document}
\maketitle

\begin{abstract}
Running large language models (LLMs) on consumer hardware is bottlenecked by the key-value (KV) cache, which grows with every decoded token and quickly exhausts the memory and bandwidth of a single consumer GPU.
We present \textbf{SparPR} (Sparse Partial Recomputation), a runtime KV-cache recovery mechanism that does not modify model weights and requires no offline training.
SparPR couples three online decisions: \emph{Query-Aware key-block selection} scores historical blocks against the current query; \emph{dynamic budget control} estimates per-block copy and recompute costs to decide how many blocks to retain; and \emph{on-demand local recomputation with cost-aware fallback} merges retained blocks into continuous spans and switches between sparse recomputation and full recovery according to a runtime cost comparison.
Implemented on FlexGen/FlexLLMGen, SparPR is evaluated on a single consumer laptop (RTX 3070 Ti 8\,GB, Ryzen 9 6900HX, 16\,GB RAM).
It cuts end-to-end latency by 31\%, 52\%, and 63\% at 1k, 2k, and 4k prompt tokens respectively, with 1.4$\times$--2.7$\times$ higher throughput, and completes an 8k-token workload in 44 minutes where the FlexGen baseline times out.
Under a fixed cache budget, SparPR retains a Needle Accuracy that stays well above StreamingLLM and H2O at every keep ratio, and ablation studies confirm that the three mechanisms act jointly.
\end{abstract}

\section{Introduction}

Deploying large language models (LLMs) on consumer hardware (laptops and single-GPU machines) is increasingly common for offline, privacy-sensitive, and cost-sensitive workloads.
The dominant cost is the key-value (KV) cache, which grows linearly with the number of tokens processed and quickly turns limited GPU memory and host--device bandwidth into first-order bottlenecks~\cite{opt,flexgen,flash}.
On datacenter GPUs these costs can be absorbed by large memory and high bandwidth; on a consumer laptop they translate into lower throughput, higher latency, and out-of-memory (OOM) failures long before the model's nominal context window is reached~\cite{streamingllm}.

Existing work addresses parts of this problem. Layered-execution and offloading methods~\cite{flexgen} place weights, activations, and caches across GPU, CPU, and disk to improve runnability. Cache-retention methods~\cite{streamingllm,h2o,arkvale,prophetkv,shadowkv} select a subset of the historical cache to keep. Local-recovery methods~\cite{quest} recompute or reload only part of the cache to avoid restoring everything. These lines of work answer \emph{whether the system can run}, \emph{which information to keep}, and \emph{how to recover it}, respectively, but they do not form a unified runtime loop that couples query relevance, budget control, and recovery-path selection.

In this paper we ask three questions that arise once a model no longer fits its KV cache in GPU memory:
(i) which historical blocks are most relevant to the current query and therefore worth keeping; (ii) how many blocks should be retained given the context length, bandwidth, and compute cost; and (iii) whether the retained blocks should be restored by local recomputation or by full recovery.
To answer them, we propose \textbf{SparPR} (Sparse Partial Recomputation), a runtime optimization method for KV-cache recovery on consumer hardware.

Our contributions are as follows.
\begin{itemize}
  \item A \textbf{Query-Aware key-block selection} mechanism that scores historical blocks by their relevance to the current query and performs Top-$k$ selection, converting uniform cache access into relevance-driven access.
  \item A \textbf{dynamic budget control} mechanism that estimates per-block copy and recompute costs and adaptively derives the target keep ratio and block count, with a conservative lower bound (\emph{Minimum Keep Floor}).
  \item An \textbf{on-demand local recomputation with cost-aware fallback} mechanism that merges discrete key blocks into continuous spans and selects between sparse and dense recovery paths by a runtime cost comparison (\emph{Dense Fallback Threshold}).
  \item A full prototype on FlexGen/FlexLLMGen, evaluated on a single consumer laptop, that cuts latency by 31\%--63\% at 1k--4k tokens and completes an 8k-token workload where the FlexGen baseline times out.
\end{itemize}

\section{Related Work}

\paragraph{Offloading and layered execution.}
FlexGen~\cite{flexgen} offloads weights and the KV cache across GPU, CPU, and disk to run large models on limited memory. LeoAM~\cite{leoam} extends this to a single commodity GPU with importance-aware hierarchical GPU--CPU--Disk KV management. These systems target runnability but do not jointly decide query relevance, budget, and recovery path.

\paragraph{KV-cache eviction and retention.}
StreamingLLM~\cite{streamingllm} keeps the initial ``attention sink'' and recent tokens, a position-based strategy. H2O~\cite{h2o} keeps heavy-hitter tokens selected by accumulated attention statistics. ArkVale~\cite{arkvale} organizes tokens into pages and recalls important evicted pages from a CPU-side backup. ShadowKV~\cite{shadowkv} offloads the value cache and reconstructs sparse KV on the fly. Most of these methods keep a budget that is fixed or decoupled from both the current query and the hardware cost.

\paragraph{Query-aware sparsity and local recomputation.}
Quest~\cite{quest} scores and loads only the most relevant page per attention head, showing that query-aware selection can be made cheap enough for inference. ProphetKV~\cite{prophetkv} reuses KV cache in retrieval-augmented generation via query-driven selective recomputation. Najafizadeh et al.~\cite{restore} study the restore-versus-recompute trade-off and switch between the two with an I/O-aware policy at the storage layer. SparPR shares the query-aware idea but works at \emph{block} granularity on an \emph{offloading} base (FlexGen) where both data movement and recomputation are expensive, and couples an explicit \emph{copy-versus-recompute cost model} with query-aware selection and a dense fallback threshold in a single runtime loop.

\section{Method}

\subsection{Overview and Problem Setting}
\label{sec:overview}

We target memory-constrained decoding on a single consumer device. At each decoding step, the system must recover (load or recompute) the historical KV cache needed to attend over the past. Let $\mathbf{q}$ be the current query representation, and let $\mathcal{M} = \{m_1, \dots, m_n\}$ be the set of $n$ candidate historical blocks, where each $m_i$ is a block-level cache unit. Let $\Omega$ denote the runtime and hardware state (context length, block size, host bandwidth, recompute efficiency, and conservative parameters).

\begin{figure}[H]
\centering
\small
\begin{tikzpicture}[
  node distance=6mm,
  proc/.style={draw=blue!55!black, fill=blue!8, rounded corners=2pt, align=center, minimum height=9mm, inner sep=5pt},
  dec/.style={draw=orange!75!black, fill=orange!12, align=center, inner sep=6pt, diamond, aspect=2.6},
  res/.style={draw=green!45!black, fill=green!8, rounded corners=2pt, align=center, minimum height=8mm, inner sep=5pt},
  arr/.style={-{Stealth}, thick}
]
\node[proc] (in) {Inputs: query $\mathbf{q}$, blocks $\mathcal{M}$, state $\Omega$};
\node[proc, below=of in] (score) {1.\ Query-Aware scoring \\ $s_i=\mathrm{Score}(\mathbf{q}, m_i)$};
\node[proc, below=of score] (budget) {2.\ Dynamic budget \\ $c_{\mathrm{copy}}$ vs.\ $c_{\mathrm{recompute}}$ $\to r_{\mathrm{keep}}, k$};
\node[proc, below=of budget] (topk) {3.\ Top-$k$ selection: $\mathcal{S}_{\mathrm{topk}}$};
\node[proc, below=of topk] (span) {4.\ Span merge: $\mathcal{S}_{\mathrm{spans}}$};
\node[dec, below=of span] (cmp) {$C_{\mathrm{sparse}} \ge \gamma\, C_{\mathrm{dense}}$ ?};
\node[res, below=of cmp, xshift=-16mm] (dense) {Dense Fallback};
\node[res, below=of cmp, xshift=16mm] (sparse) {Sparse Recompute};
\draw[arr] (in) -- (score);
\draw[arr] (score) -- (budget);
\draw[arr] (budget) -- (topk);
\draw[arr] (topk) -- (span);
\draw[arr] (span) -- (cmp);
\draw[arr] (cmp.south) -- (dense.north) node[midway, above left]{yes};
\draw[arr] (cmp.south) -- (sparse.north) node[midway, above right]{no};
\end{tikzpicture}
\caption{The SparPR runtime decision loop: query-aware scoring, dynamic budget control, span merging, and cost-aware path selection.}
\label{fig:overview}
\end{figure}
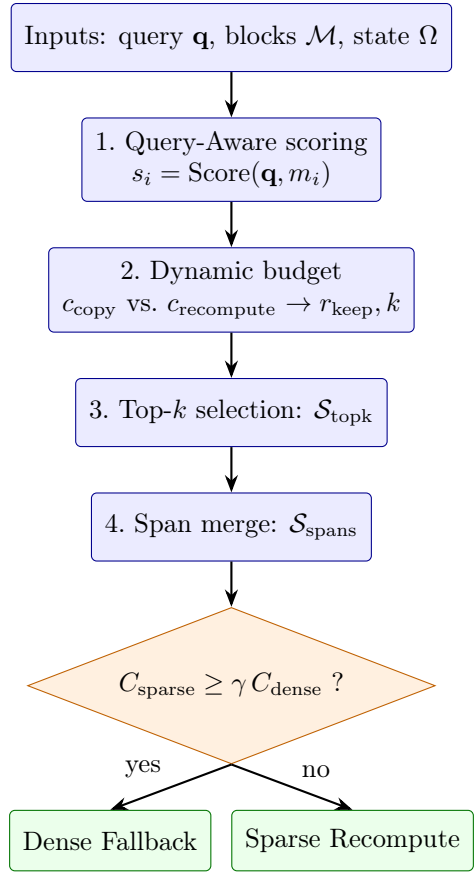

SparPR operates as an online runtime layer on top of an existing execution engine; it does not change model weights and requires no offline training. Its decision output is a recovery plan consisting of (i) a set of retained key blocks and their target count, and (ii) the recovery path (sparse recomputation or full recovery). Figure~\ref{fig:overview} summarizes the decision loop.

\subsection{Query-Aware Key-Block Selection}
\label{sec:selection}

The first stage scores each historical block by its relevance to the current query. For each block $m_i$ we precompute a compact summary, the per-head mean key vector
\begin{equation}
\bar{\mathbf{k}}_{i,h} = \frac{1}{|m_i|}\sum_{t \in m_i} \mathbf{k}_{t,h} \in \mathbb{R}^d,
\end{equation}
where $\mathbf{k}_{t,h}$ is the key vector of token $t$ at head $h$. These summaries form an $n\times H\times d$ tensor that stays resident on the GPU, so scoring touches only the summary and never the full KV blocks offloaded to CPU or disk; this is what keeps the estimate cheap. The relevance score of a block is the query--summary similarity aggregated over heads,
\begin{equation}
s_i = \sum_{h} \langle \mathbf{q}_h, \bar{\mathbf{k}}_{i,h} \rangle,
\end{equation}
a single scalar computed from $H$ dot products of length $d$; aggregating with $\max_h$ behaves similarly. Throughout, blocks, summaries, and costs are defined per layer, and the layer count $L$ is a common factor that cancels in the relative cost comparisons of Sections~\ref{sec:budget} and \ref{sec:fallback}.

Given the scores, the system selects the top-$k$ blocks:
\begin{equation}
\mathcal{S}_{\mathrm{topk}} = \mathrm{TopK}\big(\{s_i\}, k\big).
\end{equation}
Unlike a fixed threshold, Top-$k$ keeps the selection scale stable as the number of candidate blocks changes. The value of $k$ is \emph{not} a static hyperparameter; it is produced by the dynamic budget control described next.

\subsection{Dynamic Budget Control}
\label{sec:budget}

The number of blocks worth keeping depends on context length, host bandwidth, and compute cost. SparPR estimates two per-block quantities. Let each block contain $b$ tokens, use $h$ attention heads of dimension $d$, and store activations in $\beta$ bytes. The copy cost of one block is
\begin{equation}
c_{\mathrm{copy}} = \frac{2\, b\, h\, d\, \beta}{BW_{\mathrm{host}}},
\end{equation}
where $BW_{\mathrm{host}}$ is the effective host-to-device bandwidth and the factor $2$ accounts for both key and value.
The recompute cost of one block scales with the block length and the square of the hidden dimension $d_{\mathrm{hidden}}$:
\begin{equation}
c_{\mathrm{recompute}} \approx \frac{b\, d_{\mathrm{hidden}}^2}{F_{\mathrm{gpu}}},
\end{equation}
where $F_{\mathrm{gpu}}$ is an effective compute-throughput term corrected by a runtime efficiency discount. On the sparse path, a block's K/V is rebuilt from a checkpointed hidden state via the attention projection only ($\mathbf{K}=\mathbf{h}\,\mathbf{W}_K$, $\mathbf{V}=\mathbf{h}\,\mathbf{W}_V$), without re-running the MLP or a full forward pass.

Let the copy-cost ratio be
\begin{equation}
\rho_{\mathrm{copy}} = \frac{c_{\mathrm{copy}}}{c_{\mathrm{copy}} + c_{\mathrm{recompute}}}.
\end{equation}
Intuitively, when copying dominates, the system should keep more blocks to avoid repeated loads; when recomputing dominates, it should keep fewer. SparPR maps $\rho_{\mathrm{copy}}$ to a target keep ratio $r_{\mathrm{keep}}$ with a light linear mapping plus a context-length correction that adds a small retention gain in longer contexts. The target block count is then
\begin{equation}
k = \max\big(k_{\min},\, \lceil n \cdot r_{\mathrm{keep}} \rceil\big),
\end{equation}
where $k_{\min}$ implements the \emph{Minimum Keep Floor}, a conservative lower bound that prevents the budget from shrinking too aggressively under pressure.

\subsection{Span Merging and Cost-Aware Fallback}
\label{sec:fallback}

The Top-$k$ blocks are often scattered across the context. Recovering them one by one introduces fragmentation, scheduling, and fixed per-block overhead. SparPR therefore merges adjacent key blocks into continuous spans $\mathcal{S}_{\mathrm{spans}} = \{\mathrm{span}_1, \dots, \mathrm{span}_m\}$.

We model the cost of the two recovery paths. Let $n$ be the number of candidate blocks and $c_{\mathrm{copy}}$ the per-block copy cost. Full recovery costs
\begin{equation}
C_{\mathrm{dense}} = n \cdot c_{\mathrm{copy}}.
\end{equation}
Sparse recovery costs
\begin{equation}
C_{\mathrm{sparse}} = |\mathcal{S}_{\mathrm{topk}}| \cdot c_{\mathrm{recompute}} + |\mathcal{S}_{\mathrm{spans}}| \cdot c_{\mathrm{span}} + c_{\mathrm{fixed}},
\end{equation}
where $c_{\mathrm{span}}$ is the per-span organizational overhead and $c_{\mathrm{fixed}}$ is the fixed sparse-recovery overhead. Note that $C_{\mathrm{sparse}}$ depends not only on how many blocks are kept, but also on how many spans they form: more fragmented blocks raise the cost.

The path decision is governed by a \emph{Dense Fallback Threshold} $\gamma$:
\begin{equation}
\text{path} =
\begin{cases}
\text{Dense Fallback}, & \text{if } C_{\mathrm{sparse}} \ge \gamma \cdot C_{\mathrm{dense}},\\[2pt]
\text{Sparse Recompute}, & \text{otherwise}.
\end{cases}
\end{equation}
When sparse recovery is no longer cheaper than a fraction of full recovery, the system falls back to the dense path rather than persisting with an unprofitable sparse strategy. A smaller $\gamma$ makes the system fall back more readily (more conservative); a larger $\gamma$ favors sparse recovery.

\subsection{Runtime Decision Procedure}
\label{sec:algorithm}

The full decision procedure is summarized below.
\begin{center}
\begin{tabular}{l}
\toprule
\textbf{Algorithm 1}~~SparPR runtime decision.\\
\midrule
\textbf{Input:} current query $\mathbf{q}$, historical block set $\mathcal{M}$, runtime state $\Omega$\\
\textbf{Output:} recovered cache (sparse or dense)\\
\midrule
1:~~Compute block scores $s_i = \mathrm{Score}(\mathbf{q}, m_i)$ for all $m_i$\\
2:~~Estimate keep ratio $r_{\mathrm{keep}}$ from $\Omega$\\
3:~~Compute target count $k = \max(k_{\min}, \lceil n \cdot r_{\mathrm{keep}}\rceil)$\\
4:~~Select $\mathcal{S}_{\mathrm{topk}} = \mathrm{TopK}(\{s_i\}, k)$\\
5:~~Merge adjacent blocks into spans $\mathcal{S}_{\mathrm{spans}}$\\
6:~~Estimate $C_{\mathrm{dense}}$ and $C_{\mathrm{sparse}}$\\
7:~~\textbf{if} $C_{\mathrm{sparse}} \ge \gamma \cdot C_{\mathrm{dense}}$ \textbf{then} Dense Fallback\\
8:~~\textbf{else} Sparse Recompute\\
\bottomrule
\end{tabular}
\end{center}
\section{Experiments}

\subsection{Setup}
\label{sec:setup}

\textbf{Hardware and software.} All results are real runs on a single consumer laptop with an NVIDIA RTX 3070 Ti (8\,GB), an AMD Ryzen 9 6900HX mobile CPU, 16\,GB host memory, and a local SSD, running Windows with WSL. The execution engine is FlexGen/FlexLLMGen. We evaluate \texttt{facebook/opt-6.7b}~\cite{opt} with batch size 4 and a static offloading policy, a setting in which the KV cache quickly stresses the 8\,GB GPU budget.

\textbf{Baselines.} We compare against \emph{FlexGen}~\cite{flexgen} (system runnability baseline), \emph{StreamingLLM}~\cite{streamingllm} (position-based retention) and \emph{H2O}~\cite{h2o} (statistics-based retention) for needle-style retrieval fidelity, and \emph{Quest}-style query-aware selection for the ablation.

\textbf{Needle task.} For needle-style retrieval fidelity we use a constructed needle task: a target fact (``The Wi-Fi password for the guest network at the SWJTU is 2022116822'') is inserted at one of two depth ratios, 0.15 or 0.85 of the context, into a long distractor text (an excerpt of \emph{Alice in Wonderland}), and the model is asked ``What is the Wi-Fi password for the guest network at the SWJTU?''. Needle Accuracy is the hit rate averaged over the two insertion depths.

\subsection{System Performance and Runnability}
\label{sec:perf}

Table~\ref{tab:perf} reports end-to-end latency and throughput as the prompt length grows from 1k to 8k tokens. DeepSpeed and HF Accelerate run out of memory before 1k, and a Quest-style query-aware selection OOMs at 1200 tokens; only the offloading FlexGen baseline and SparPR keep a runnable range. FlexGen completes 1k--4k by spilling to CPU and disk but slows to the point of timing out at 8k. SparPR completes every length, cutting latency by 31\%, 52\%, and 63\% at 1k, 2k, and 4k tokens and reaching 1.4$\times$, 2.1$\times$, and 2.7$\times$ higher throughput; the gap widens with context length because SparPR bounds the growth of the KV-cache recovery cost.

\begin{table}[H]
\centering
\caption{End-to-end latency (s) and throughput (tok/s) of FlexGen and SparPR at increasing prompt length (OPT-6.7B, batch 4, RTX 3070 Ti 8\,GB). ``timeout'' denotes failure to complete within the experiment's time budget.}
\label{tab:perf}
\small
\begin{tabular}{lrrrrr}
\toprule
Prompt & FlexGen lat. & FlexGen thr. & SparPR lat. & SparPR thr. & Lat. reduction \\
\midrule
1k & 1064 & .466 & 737 & .673 & 31\% \\
2k & 2101 & .236 & 1009 & .491 & 52\% \\
4k & 4174 & .119 & 1553 & .319 & 63\% \\
8k & timeout & --- & 2641 & .187 & --- \\
\bottomrule
\end{tabular}
\end{table}

\subsection{Needle-Style Retrieval Fidelity Under Budget}
\label{sec:fidelity}

Table~\ref{tab:fidelity} reports Needle Accuracy at fixed cache keep ratios across 2k, 4k, and 8k prompt tokens. At every keep ratio and length, SparPR stays well above H2O and StreamingLLM: at high budgets (30--50\%) it retains .60--.94 accuracy versus .31--.51 for H2O and .15--.49 for StreamingLLM, and at 10\% keep it retains .60/.48/.31 at 2k/4k/8k while H2O falls to around .12 and StreamingLLM collapses to .02. StreamingLLM is position-based, so its needle survives only if it falls in the recent window; H2O is query-agnostic, so its hit rate roughly tracks the keep ratio and is nearly length-invariant. SparPR's query-aware selection keeps the needle-relevant block even at low budgets (at the cost of modestly higher latency), but its accuracy still degrades with context length, since longer distractors make the needle-relevant block harder to identify.

\begin{table}[H]
\centering
\caption{Needle Accuracy at fixed cache keep ratios, at 2k/4k/8k prompt tokens (mean of three runs).}
\label{tab:fidelity}
\small
\begin{tabular}{llccc}
\toprule
Method & Keep ratio & 2k & 4k & 8k \\
\midrule
StreamingLLM & 50\% & .445 & .467 & .485 \\
 & 40\% & .291 & .333 & .345 \\
 & 30\% & .153 & .197 & .208 \\
 & 20\% & .020 & .051 & .070 \\
 & 10\% & .020 & .020 & .020 \\
 & 5\% & .020 & .020 & .020 \\
\midrule
H2O & 50\% & .492 & .506 & .511 \\
 & 40\% & .408 & .393 & .408 \\
 & 30\% & .321 & .308 & .313 \\
 & 20\% & .212 & .221 & .215 \\
 & 10\% & .120 & .127 & .114 \\
 & 5\% & .065 & .069 & .073 \\
\midrule
SparPR & 50\% & .943 & .879 & .780 \\
 & 40\% & .898 & .831 & .694 \\
 & 30\% & .847 & .772 & .598 \\
 & 20\% & .763 & .654 & .473 \\
 & 10\% & .600 & .482 & .308 \\
 & 5\% & .424 & .330 & .208 \\
\bottomrule
\end{tabular}
\end{table}

\subsection{Parameter Analysis}
\label{sec:params}

Table~\ref{tab:floor} shows the effect of the Minimum Keep Floor. Lowering the floor reduces the average keep ratio and the sparse cost and raises throughput, while accuracy degrades only gradually---consistent with the query-aware selection retaining the more relevant blocks first.

\begin{table}[H]
\centering
\caption{Effect of Minimum Keep Floor on keep ratio, throughput, accuracy, and sparse cost (needle task).}
\label{tab:floor}
\small
\begin{tabular}{lcccc}
\toprule
Floor & Avg. keep & Throughput & Needle acc. & Sparse cost (ms) \\
\midrule
1.00 & 100.0\% & 0.482 & 0.993 & 952.33 \\
0.90 & 90.0\% & 0.532 & 0.987 & 891.67 \\
0.80 & 80.0\% & 0.588 & 0.983 & 821.33 \\
0.75 & 75.5\% & 0.642 & 0.973 & 761.67 \\
0.70 & 72.6\% & 0.681 & 0.963 & 712.33 \\
0.65 & 69.7\% & 0.728 & 0.947 & 671.67 \\
0.60 & 66.8\% & 0.784 & 0.927 & 632.33 \\
\bottomrule
\end{tabular}
\end{table}

Table~\ref{tab:gamma} shows the effect of the Dense Fallback Threshold. A lower threshold makes the system fall back to full recovery more often (higher bypass count) and behave more conservatively; a higher threshold favors sparse recovery, raising throughput while accuracy declines slightly.

\begin{table}[H]
\centering
\caption{Effect of Dense Fallback Threshold on throughput, accuracy, and bypass count (needle task).}
\label{tab:gamma}
\small
\begin{tabular}{lccc}
\toprule
Threshold $\gamma$ & Throughput & Needle acc. & Bypass count \\
\midrule
0.72 & 0.520 & 0.985 & 12.00 \\
0.74 & 0.560 & 0.983 & 10.00 \\
0.76 & 0.610 & 0.980 & 8.00 \\
0.78 & 0.650 & 0.977 & 6.00 \\
0.80 & 0.680 & 0.971 & 5.00 \\
0.85 & 0.700 & 0.962 & 3.00 \\
0.90 & 0.710 & 0.943 & 1.00 \\
\bottomrule
\end{tabular}
\end{table}

\subsection{Ablation Study}
\label{sec:ablation}

Ablation studies use the same needle task at a fixed 1000-token prompt length, and every reported value is the mean of three runs. Note that the needle task is a different workload from the throughput sweep in Table~\ref{tab:perf}, so the FlexGen baseline latency here differs slightly from Table~\ref{tab:perf}. Tables~\ref{tab:ablation_ext} and \ref{tab:ablation_rt} isolate the contribution of each component. \emph{Quest-only} keeps query-aware selection but drops budget and fallback, improving latency but hurting accuracy. \emph{w/o Budget} keeps too much context and gains little efficiency. \emph{w/o Span Merge} incurs the highest sparse cost (1085.33\,ms) because of fragmented recovery. \emph{w/o Bypass} cannot retreat from unprofitable sparse paths and loses accuracy. \emph{SparPR-full} achieves the lowest latency and highest throughput while keeping high accuracy, confirming that the mechanisms act jointly.

\begin{table}[H]
\centering
\caption{Ablation: external results (needle task, 1000 tokens, mean of three runs).}
\label{tab:ablation_ext}
\small
\begin{tabular}{lccc}
\toprule
Method & Latency (s) & Throughput & Needle acc. \\
\midrule
FlexGen & 991.33 & 0.50 & 0.983 \\
Quest-only & 845.33 & 0.59 & 0.777 \\
w/o Budget & 838.67 & 0.60 & 0.963 \\
w/o Span Merge & 936.67 & 0.53 & 0.973 \\
w/o Bypass & 812.33 & 0.62 & 0.917 \\
SparPR-full & 756.67 & 0.66 & 0.957 \\
\bottomrule
\end{tabular}
\end{table}

\begin{table}[H]
\centering
\caption{Ablation: runtime statistics (needle task, 1000 tokens, mean of three runs).}
\label{tab:ablation_rt}
\small
\begin{tabular}{lccccc}
\toprule
Method & Avg. keep & Bypass & Blocks skip & Recompute ok & Sparse cost (ms) \\
\midrule
FlexGen & 100.0\% & 0.00 & 0.00 & 0.00 & 952.33 \\
Quest-only & 58.7\% & 0.00 & 726.67 & 3.00 & 805.67 \\
w/o Budget & 85.0\% & 3.33 & 266.33 & 4.33 & 818.33 \\
w/o Span Merge & 73.5\% & 16.33 & 468.33 & 8.67 & 1085.33 \\
w/o Bypass & 73.8\% & 0.00 & 462.67 & 12.33 & 821.67 \\
SparPR-full & 73.2\% & 5.67 & 473.33 & 9.33 & 728.33 \\
\bottomrule
\end{tabular}
\end{table}

\section{Conclusion and Limitations}

We presented SparPR, a runtime KV-cache recovery mechanism for resource-constrained LLM inference on consumer hardware. SparPR couples query-aware key-block selection, dynamic budget control, and cost-aware fallback into one online decision loop. On a single consumer laptop, it cuts latency by 31\%--63\% and raises throughput by 1.4--2.7$\times$ over FlexGen at 1k--4k tokens, completes an 8k-token workload where FlexGen times out, and keeps consistently higher needle accuracy than StreamingLLM and H2O at every cache budget; ablation confirms the three mechanisms act jointly.

Our study has limitations. Experiments use a single OPT-6.7B model, one consumer GPU, and a constructed needle task; scaling to larger models, standard long-context benchmarks, and more hardware remains future work. Finer-grained page-level cache organization and online calibration of the cost models are promising directions.

\end{document}